\documentclass[fleqn,10pt]{wlscirep}
\usepackage[utf8]{inputenc}
\usepackage[T1]{fontenc}
\usepackage{lineno}
\usepackage[numbers]{natbib}
\usepackage{subfiles}
\usepackage{graphicx}
\usepackage{float}
\usepackage{hyperref}
\usepackage{mathtools}
\usepackage{changepage}
\usepackage{caption}
\usepackage{multirow}
\usepackage{titlesec}
\usepackage{amsmath}

\usepackage{makecell}

\usepackage[math]{cellspace}
\usepackage{tikz}
\usepackage{url}

\usepackage{colortbl}
\linenumbers
\nolinenumbers

\title{The Leaf Clinical Trials Corpus: a new resource for query generation from clinical trial eligibility criteria}

\author[1,*]{Nicholas J Dobbins}
\author[2]{Tony Mullen}
\author[3]{Özlem Uzuner}
\author[1]{Meliha Yetisgen}
\affil[1]{Department of Biomedical Informatics \& Medical Education, University of Washington, Seattle, WA, USA}
\affil[2]{Khoury College of Computer Science, Northeastern University, Seattle, WA, USA}
\affil[3]{Department of Information Sciences and Technology, George Mason University, Fairfax, VA, USA}
\affil[*]{corresponding author(s): (ndobb@uw.edu)}

\begin{abstract}
\noindent
Identifying cohorts of patients based on eligibility criteria such as medical conditions, procedures, and medication use is critical to recruitment for clinical trials. Such criteria are often most naturally described in free-text, using language familiar to clinicians and researchers. In order to identify potential participants at scale, these criteria must first be translated into queries on clinical databases, which can be labor-intensive and error-prone. Natural language processing (NLP) methods offer a potential means of such conversion into database queries automatically. However they must first be trained and evaluated using corpora which capture clinical trials criteria in sufficient detail. In this paper, we introduce the Leaf Clinical Trials (LCT) corpus, a human-annotated corpus of over 1,000 clinical trial eligibility criteria descriptions using highly granular structured labels capturing a range of biomedical phenomena. We provide details of our schema, annotation process, corpus quality, and statistics. Additionally, we present baseline information extraction results on this corpus as benchmarks for future work.
\end{abstract}
\begin{document}
\flushbottom
\maketitle

\section*{Background \& Summary}
\label{sec:background_and_summary}
\noindent Randomized controlled trials serve a critical role in the generation of medical evidence and furthering of biomedical research. In order to identify patients for clinical trials, investigators publish eligibility criteria, such as past history of certain conditions, treatments, or laboratory tests. These eligibility criteria are typically composed in free-text, consisting of inclusion and exclusion criteria. Patients meeting a trial's eligibility criteria are considered potential candidates for recruitment. \\

\noindent Recruitment of participants remains a major barrier to successful trial completion \cite{richesson2013electronic}, so generating a large pool of potential candidates is often necessary. Manual chart review of hundreds or thousands of patients to determine a candidate pool, however, can be prohibitively labor- and time-intensive. Cohort discovery tools such as Leaf \cite{dobbins2019leaf} and i2b2 \cite{murphy2010serving} may be used, providing researchers with a relatively simple drag-and-drop graphical interface in their web browser to create database queries to find potential patients in electronic health records (EHR). Learning how to use such tools nevertheless presents a challenge, as graphically-represented concepts may not align with researchers' understanding of biomedical phenomena or trial eligibility criteria. In addition, certain complex queries may simply be impossible to execute due to structural limitations on the types of possible queries presented in these tools, such as complex temporal or nested sequences of events. \\

\noindent An alternative approach which holds promise is the use of natural language processing (NLP) to automatically analyze eligibility criteria and generate database queries to find patients in EHRs. NLP-based approaches have the advantage of obviating potential learning curves of tools such as Leaf, while leveraging existing eligibility criteria composed in a format researchers are already familiar with. Recent efforts to explore NLP-based approaches to eligibility criteria query generation have been published. These approaches can be generally categorized as (1) modular, multi-step methods which transform eligibility criteria into intermediate representations and finally into queries using rules \cite{yuan2019criteria2query, wang2019translate, yu2020}, (2) direct text-to-query generation methods using neural network-based semantic parsing \cite{wang2019translate, yu2020}, and (3) information retrieval approaches to detect relevant sections of free-text notes meeting a given criteria \cite{koopman2016test, liu2020implementation, park2021framework, truong2022ittc}. \\

\noindent For each category of NLP approaches, a key element for accelerating research efforts is large, robust corpora which capture eligibility criteria semantics sufficiently for high-accuracy query generation. Such corpora can serve as reliable benchmarks for purposes of comparing NLP methods as well as training datasets. A number of corpora designed for multi-step methods, which we focus on here, have been published. Past corpora cover only a modest number of eligibility criteria \cite{weng2011elixr}, are narrowly focused on certain diseases only \cite{kang2017eliie}, are not publicly available \cite{tu2011, milian2015enhancing}, or have annotations insufficiently granular to fully capture the diverse, nuanced semantics of eligibility criteria \cite{kury2020chia}. Yu \textit{et al} \cite{yu2020} released a corpus designed for direct text-to-query generation with semantic parsing, however given the relative simplicity of generated queries to date compared to the complexity of clinical databases, it's not clear this approach is yet viable for real-world clinical trials recruitment.  \\

\noindent In this paper, we present the Leaf Clinical Trials (LCT) corpus. To the best of our knowledge, the LCT corpus is the largest and most comprehensive human-annotated corpus of publicly available clinical trials eligibility criteria. The corpus is designed to accurately capture a wide range of complex, nuanced biomedical phenomena found in eligibility criteria using a rich, granular annotation schema. As the LCT annotation schema is uniquely large, fine-grained and task-oriented, the corpus can serve as a valuable training dataset for NLP approaches while significantly simplifying disambiguation steps and text-processing for query generation. The LCT annotation schema builds upon the foundational work of EliIE \cite{kang2017eliie}, an Information Extraction (IE) system for eligibility criteria, and Chia \cite{kury2020chia}, a large corpus of clinical trials of various disease domains. Expanding the EliIE and Chia annotation schemas, we developed the LCT annotation schema to greatly increase the variety of biomedical phenomena captured while also annotating eligibility criteria semantics at a significantly more granular level. Table \ref{tbl_corpora_compare} presents a comparison of the LCT corpus and these corpora. \\

\noindent In the following sections, we (1) discuss the LCT corpus annotation schema, (2) include descriptive statistics on corpus structure, (3) provide baseline named entity recognition and relation extraction performance using the corpus, and (4) discuss areas of future potential for query generation in the Usage Notes section.

\section*{Methods}
\label{sec:methods}
\begin{table}
\centering
\begin{tabular}{l|lll} 
 \toprule
 Measure & EliIE \cite{kang2017eliie} & Chia \cite{kury2020chia} & \textbf{LCT Corpus} \\
 \hline
    Disease domain & Alzheimer's Disease & All & \textbf{All} \\
    No. of Eligibility Descriptions & 230 & 1,000 & \textbf{1,006} \\
    No. of Annotations & 15,596 & 68,174 & \textbf{105,816} \\
    No. of Entity types & 8 & 15 & \textbf{50} \\
    No. of Relation types & 3 & 12 & \textbf{51} \\
    Mean Entities per doc. & - & 46 & \textbf{105} \\
    Mean Relations per doc. & - & 19 & \textbf{49} \\
 \hline
\end{tabular}
\caption{\textbf{Annotation statistics for EliIE, Chia, and LCT corpora.}}
\label{tbl_corpora_compare}
\end{table}

\subsection*{Eligibility Criteria and Database Queries}
\begin{figure}[t]
  \includegraphics[scale=0.57]{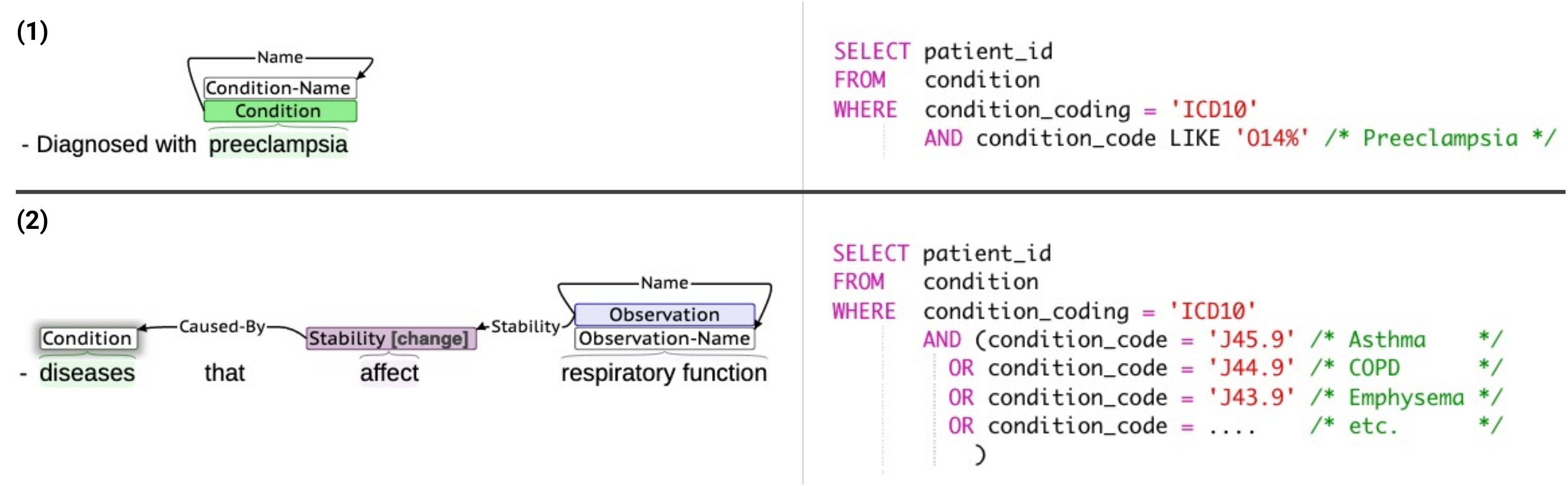}  
\caption{Example eligibility criteria annotated used the LCT corpus annotation schema (left) and corresponding example SQL queries (right) using a hypothetical database table and columns. Annotations were done using the Brat annotation tool \cite{stenetorp2012brat}. The ICD-10 codes shown are examples and not intended to be exhaustive.}
\label{fig_lct_text2sql}
\end{figure}

\noindent The NLP tasks involved in transforming eligibility criteria into database queries include \textbf{named entity recognition} (NER) to tag meaningful spans of text as named entities, \textbf{relation extraction} to classify relations between named entities, \textbf{normalization} to map named entities to common coded representations (e.g., ICD-10), \textbf{negation detection} to detected negated statements (e.g., "not hypertensive") and so on. Gold standard corpora quality can thus directly affect performance and the validation of each of these tasks. In this article, we only focus on design and development of the LCT corpus. Figure \ref{fig_lct_text2sql} illustrates why corpora structure and integrity are important for the task of query generation, using examples of eligibility criteria annotated using the LCT annotation schema and corresponding hypothetical Structured Query Language (SQL) queries. In the first eligibility criterion, "preeclampsia" is explicitly named, and thus can be directly normalized to an International Classification of Diseases-Tenth Revision (ICD-10) or other coded representation. However, eligibility criteria involving non-specific drugs, conditions, procedures, contraindications, and so on are used frequently in clinical trials. In the second criterion in Figure \ref{fig_lct_text2sql}, "diseases" in "diseases that affect respiratory function" is non-specific, and must be reasoned upon in order to determine appropriate codes, such as asthma, chronic obstructive pulmonary disease (COPD), or emphysema.  Programmatically reasoning to generate queries in such cases would be challenging and often impossible if the underlying semantics were not captured appropriately. With this in mind, we developed the LCT annotation schema in order to enable reasoning and ease query generation for real-world clinical trials use. As the second example in Figure \ref{fig_lct_text2sql} shows, the LCT annotation captures the semantics of complex criteria, with changes to "respiratory function" annotated using a \textit{Stability[change]} entity and \textit{Stability} relation, and the cause, "diseases" annotated with a \textit{Caused-By} relation. During query generation, a hypothetical algorithm can thus use LCT entities and relations to first normalize the span "respiratory function", then reason that asthma, COPD, emphysema, and other conditions can affect respiratory function and thus the generated query should find patients with those diagnoses.

\subsection*{Annotation schema}

\noindent We aimed to develop an expressive, task-oriented annotation schema which could capture a wide range of medical concepts and logical constructs present in eligibility criteria. To accomplish this, we first analyzed previously published corpora \cite{weng2011elixr,boland2012elixrtime,kang2017eliie,kury2020chia} and expanded the list of included biomedical phenomena to fully capture the context and logic present in real clinical trials criteria. As one example, we introduced an entity called \textit{Contraindication} to reflect where use of a given treatment is inadvisable due to possible harm to the patient. \\

\noindent The LCT annotation schema is designed with the following goals and assumptions:

\begin{enumerate}
    \item The annotation schema should be \textbf{practical} and  \textbf{task-oriented} with a focus on facilitating ease of query generation. 
    \item A greater number of \textbf{more specific}, \textbf{less ambiguous} annotated phenomena should be favored over a smaller number of possibly ambiguous ones.
    \item Annotations should be \textbf{easily transformable} into composable, interconnected programmatic objects, trees, or node-edge graph representations.
    \item The annotation schema should \textbf{model eligibility criteria intent and semantics} as closely as possible in order to ensure generated queries can do the same.
\end{enumerate}

\noindent The LCT annotation schema is composed of \textbf{entities} and 
\textbf{relations}. Entities refer to biomedical, demographic, or other named entities relevant to eligibility criteria, and are annotated as a span of one or more tokens. We organized LCT entities into the following categories:
\begin{itemize}
    \item \textbf{Clinical} - \textit{Allergy, Condition, Condition-Type, Code, Contraindication, Drug, Encounter, Indication, Immunization, Observation, Organism, Specimen, Procedure, Provider}. 
    \item \textbf{Demographic} - \textit{Age, Birth, Death, Ethnicity, Family-Member, Language, Life-Stage-And-Gender}. 
    \item \textbf{Logical} - \textit{Exception, Negation}. 
    \item \textbf{Qualifiers} - \textit{Acuteness, Assertion, Modifier, Polarity, Risk, Severity, Stability}. 
    \item \textbf{Temporal and Comparative} - \textit{Criteria-Count, Eq-Comparison} (an abbreviation of "Equality Comparison"), \textit{Eq-Operator, Eq-Temporal-Period, Eq-Temporal-Recency, Eq-Temporal-Unit, Eq-Unit, Eq-Value}. 
    \item \textbf{Other} - \textit{Coreference, Insurance, Location, Other, Study}. 
\end{itemize}

\noindent The LCT corpus also includes 7 \textit{Name} entities: \textit{Allergy-Name}, \textit{Condition-Name}, \textit{Drug-Name}, \textit{Immunization-Name}, \textit{Observation-Name}, \textit{Organism-Name} and \textit{Procedure-Name}. \textit{Name} entities serve a special purpose in the LCT corpus, as they indicate that a span of text refers to a \textit{specific} condition, drug, etc., as opposed to \textit{any} condition or drug. \textit{Name} entities overlap with their respective general entities. For example, the span "preeclampsia" refers to a specific condition, and would thus be annotated as both a \textit{Condition} and \textit{Condition-Name}, while the span "diseases" is non-specific and would be annotated as only \textit{Condition}. A full listing of the LCT annotation guidelines can be found at \url{https://github.com/uw-bionlp/clinical-trials-gov-annotation/wiki}. \\

\noindent We defined a total of 50 entities in the LCT corpus. Examples of selected representative entities are presented in Table \ref{tbl_entity_examples}. In our representation, a subset of entities have \textbf{values} as well. For example, an \textit{Encounter} may have a value of \textit{emergency}, \textit{outpatient} or \textit{inpatient}. Values are optional in some entities (such as \textit{Encounters} or \textit{Family-Member}, where they may not always be clear or are intentionally broad) and always present in others. In the example annotations presented below, values are denoted using brackets ("[...]") following entity labels. \\

\begin{table}
    \def\arraystretch{1.3}
\begin{tabular}{m{2cm} m{2.5cm} m{4.9cm} m{6.5cm}}
    \toprule
    \textbf{Category} & \textbf{Entity} & \textbf{Values} & \textbf{Example Text}\\
    \hline 
    &
        Condition 
            & --  
            & Diagnosed with $\underset{Condition}{\underline{\mathrm{hypertension}}}$ in past year \\
     & Contraindication
            & --
            & any $\underset{Contraindication}{\underline{\mathrm{contraindications}}}$ to vaginal delivery \\
     & Drug 
            & -- 
            & on $\underset{Drug}{\underline{\mathrm{beta\ blockers}}}$ \\
     Clinical & Encounter 
            & emergency, outpatient, inpatient 
            & recently $\underset{Encounter[inpatient]}{\underline{\mathrm{admitted}}}$ to a hospital \\
     & Immunization 
            & --
            & received $\underset{Immunization}{\underline{\mathrm{Influenza\ vaccination}}}$ \\
     & Observation
            & lab, vital, clinical-score, survey, social-habit
            & $\underset{Observation[lab]}{\underline{\mathrm{Platelet\ count}}}$ less than 500 \\
     & Procedure
            & -- 
            & Undergoing or scheduled for a $\underset{Procedure}{\underline{\mathrm{colonoscopy}}}$ \\[2ex]
    \hline       
    \multirow{4}{*}[-8pt]{\mbox{Demographic}} & 
        Age 
            & -- 
            & 43 years $\underset{Age}{\underline{\mathrm{old}}}$ \\
     & Birth
            & --
            & $\underset{Birth}{\underline{\mathrm{born}}}$ within the past 6 months \\
     & Family-Member 
            & mother, father, sibling, etc.
            & history of $\underset{Family-Member[mother]}{\underline{\mathrm{maternal}}}$ breast cancer \\
     & Language 
            & -- 
            & Speaks $\underset{Language}{\underline{\mathrm{English}}}$ or $\underset{Language}{\underline{\mathrm{Spanish}}}$  \\[2ex]
    \hline
    Logical &
       Negation 
            & -- 
            & with $\underset{Negation}{\underline{\mathrm{no}}}$ systemic disease \\
            
    \hline
    
    \multirow{6}{*}[-16pt]{\mbox{Qualifier}} &
        Assertion
            & intention, hypothetical, possible
            & which $\underset{Assertion[hypothetical]}{\underline{\mathrm{may}}}$ cause conditions \\
         & Modifier
            & --
            & $\underset{Modifier}{\underline{\mathrm{alcohol}}}$ or $\underset{Modifier}{\underline{\mathrm{substance}}}$ abuse  \\
         & Polarity 
                & low, high, positive, negative 
                & showing $\underset{Polarity[high]}{\underline{\mathrm{elevated}}}$ serum creatinine  \\
         & Risk
            & -- 
            & at heightened $\underset{Risk}{\underline{\mathrm{potential}}}$ for suicide \\
         & Severity 
            & mild, moderate, severe 
            & with $\underset{Severity[severe]}{\underline{\mathrm{serious}}}$ complications from surgery  \\
         & Stability 
                & stable, change 
                & conditions known to $\underset{Stability[change]}{\underline{\mathrm{affect}}}$ mood \\[2ex]
    \hline
    \multirow{4}{*}[-17pt]{\mbox{Temporal and} Comparative} &
        Criteria-Count
            & --
            & at least 3 of $\underset{Criteria-Count}{\underline{\mathrm{the\ following\ conditions}}}$: \\
     & Eq-Comparison
            & --
            & $\underset{Eq-Comparison}{\underline{\mathrm{greater\ than\ 50ml}}}$ \\
     
     & Eq-Temporal-Period
            & past, present, future
            & $\underset{Eq-Temporal-Period[present]}{\underline{\mathrm{Active}}}$ illness \\
     & Eq-Temporal-Recency
            & first-time, most-recent
            & $\underset{Eq-Temporal-Recency[most-recent]}{\underline{\mathrm{Latest}}}$ BMI > 35 \\
            
    \hline   
    Other &
       Location 
            & residence, clinic, hospital, unit, emergency-department  
            & Seen at $\underset{Location[clinic]}{\underline{\mathrm{diabetes\ care\ clinic}}}$ \\[2ex]
    
\end{tabular}
    \caption{\textbf{Examples of representative LCT annotation schema entities.} A full listing of all entities can be found in the LCT annotation guidelines at \url{https://github.com/uw-bionlp/clinical-trials-gov-annotation/wiki}.}
    \label{tbl_entity_examples}
\end{table} 

\noindent Relations serve as semantically meaningful connections between entities, such as when one entity acts upon, is found by, caused by, or related in some way to another. We categorize relations into the following:
\begin{itemize}
    \item \textbf{Alternatives and Examples} - \textit{Abbrev-Of, Equivalent-To, Example-Of}. 
    \item \textbf{Clinical} - \textit{Code, Contraindicates, Indication-For, Name, Provider, Specimen, Stage, Type}. 
    \item \textbf{Dependent} - \textit{Caused-By, Found-By, Treatment-For, Using}. 
    \item \textbf{Logical} - \textit{And, If-Then, Negates, Or}. 
    \item \textbf{Qualifier} - \textit{Acuteness, Asserted, Dose, Modifies, Polarity, Risk-For, Severity, Stability}.
    \item \textbf{Temporal and Comparative} - \textit{After, Before, Criteria, Duration, During, Max-Value, Min-Value, Minimum-Count, Numeric-Filter, Operator, Per, Temporal-Period, Temporal-Recency, Temporal-Unit, Temporality, Unit, Value}.
    \item \textbf{Other} - \textit{From, Except, Has, Is-Other, Location, Refers-To, Study-Of}.
\end{itemize}

\noindent We defined a total of 51 relations in the LCT corpus. Examples of relation criteria are shown in Table \ref{tbl_relation_examples}.  \\

\noindent In our annotations, some entity spans overlap with other entity spans in order to fully capture complex underlying semantics. Consider for example, the expression "Ages 18-55 years old". While an \textit{Age} entity may be assigned to token "Ages", if an \textit{Eq-Comparison} entity alone were assigned to the span "18-55 years old", the underlying semantics of the tokens "18", "-", "55", and "years" would be lost. In the following examples, we use the term \textbf{fine-grained entity} to refer to entities which are sub-spans of other \textbf{general entities}. Fine-grained entities are linked to general entities by relations. We use down arrow symbols (↓) to denote entity annotation and left and right arrow symbols (← and →) to denote relations. The (+) symbols denote overlapping entities on the same span. \\

\noindent The example expression "Ages 18-55 years old" would be annotated in three layers. In the first layer, the expression is annotated with \textit{Age} and \textit{Eq-Comparison} general entities with a relation between them: \\

\begin{center}
\begin{tabular}{c c c c c c c}
    "Ages" & & \multicolumn{5}{c}{$\underbrace{\text{"18 \, \, - \, \, 55 \, \, years \, \, old"}}$} \\ 
    \big\downarrow & & \multicolumn{5}{c}{\big\downarrow}  \\
    \textit{Age} & $\xrightarrow[Numeric-Filter]{}$ & \multicolumn{5}{c}{$\textit{Eq-Comparison}$} \\ \\
\end{tabular}
\end{center}

\noindent In the second layer, fine-grained entities with respective values are annotated: \\

\begin{center}
\begin{tabular}{c c c c}
    "18" & "-" & "55" & "years" \\ 
    \big\downarrow & \big\downarrow & \big\downarrow & \big\downarrow \\
    \textit{Eq-Value} & \textit{Eq-Operator} & \textit{Eq-Value} & \textit{Eq-Temporal-Unit} \\
     & \textit{[between]} & & \textit{[year]} \\
\end{tabular}
\end{center}

\noindent In the third layer, relations connecting fine-grained entities to the general \textit{Eq-Comparison} entity are added: \\

\begin{center}
\begin{tabular}{llll}
    \multirow{4}{5.5em}[-8pt]{\textit{\mbox{Eq-Comparison}}} & \multirow{4}{1em}[-4pt]{$\begin{cases}\\\\\\\\\end{cases}$} & $\xrightarrow[Value]{}$ & \textit{Eq-Value "18"} \\
    & & $\xrightarrow[Operator]{}$ & \textit{Eq-Operator[between]} "-" \\
    & & $\xrightarrow[Value]{}$ & \textit{Eq-Value "55"} \\
    & & $\xrightarrow[Temporal-Unit]{}$ & \textit{Eq-Temporal-Unit[year]} "years" \\
\end{tabular}
\end{center}

\noindent This multilayered annotation strategy allows significant flexibility in capturing entities and relations in a slot-filling fashion, simplifying the task of downstream query generation. We show examples of this in the Usage Notes section. \\

\begin{table*}[ht!]
    \centering
    \def\arraystretch{1.6}
\begin{tabular}{m{3.8cm} m{2.2cm} m{10cm}}
\toprule
    \textbf{Category} & \textbf{Relation} & \textbf{Example Annotation} \\ \midrule
    
     & Abbrev-Of     & $\underset{Condition}{\underline{\mathrm{Post\ Concussion\ Syndrome}}}$ \quad $\xleftarrow[Abbrev-Of]{}$ \quad ($\underset{Condition}{\underline{\mathrm{PCS}}}$) \\
        
    Alternatives and Examples & Equivalent-To & $\underset{Condition}{\underline{\mathrm{Thrombocytopenia}}}$ \quad $\xleftarrow[Equivalent-To]{}$ \quad $\underset{Observation[lab]}{\underline{\mathrm{platelets}}}$ \quad < 100,000/mm3" \\
    
     & Example-Of    & $\underset{Condition}{\underline{\mathrm{skin\ condition}}}$ \quad $\xleftarrow[Example-Of]{}$ \quad (e.g. \quad $\underset{Condition}{\underline{\mathrm{eczema}}}$ )" \\[2ex] 
     
    \hline
     
     Clinical & Contraindicates & conditions \quad  $\underset{Contraindication}{\underline{\mathrm{contraindicating}}}$ \quad $\xrightarrow[Contraindicates]{}$ \quad $\underset{Procedure}{\underline{\mathrm{MRI}}}$ \\[2ex] 
    
    \hline
    
    \multirow{4}{*}[-13pt]{\mbox{Dependent}} &
        Caused-By     & $\underset{Observation}{\underline{\mathrm{swellings}}}$ \quad $\xrightarrow[Caused-By]{}$ \quad $\mathrm{due\ to}$ \quad $\underset{Condition}{\underline{\mathrm{trauma}}}$ \\    
        
     & Found-By      & $\underset{Observation}{\underline{\mathrm{lesion}}}$ \quad $\xrightarrow[Found-By]{}$ \quad $\mathrm{seen\ on\ standard}$ \quad $\underset{Procedure}{\underline{\mathrm{imaging}}}$ \\
    
     & Treatment-For & $\underset{Procedure}{\underline{\mathrm{coronary\ bypass\ surgery}}}$ \quad $\xrightarrow[Treatment-For]{}$ \quad $\mathrm{for}$ \quad $\underset{Condition}{\underline{\mathrm{atherosclerosis}}}$ \\
    
     & Using         & $\underset{Procedure}{\underline{\mathrm{total\ knee\ arthroplasty}}}$ \quad $\xrightarrow[Using]{}$ \quad $\mathrm{with}$ \quad  $\underset{Procedure}{\underline{\mathrm{spinal\ anesthesia}}}$ \\[2ex]
    
    \hline
    
    Logical &
        If-Then       & BMI \quad $\underset{Eq-Comparison}{\underline{\mathrm{greater\ than\ 38}}}$ \quad $\xleftarrow[If-Then]{}$ \quad $\mathrm{for}$ \quad $\underset{Life-Stage-And-Gender[female]}{\underline{\mathrm{women}}}$ \\[2ex]
         
    \hline
    
         & Risk-For & $\underset{Risk}{\underline{\mathrm{risk}}}$ \quad $\xrightarrow[Risk-For]{}$ \quad $\mathrm{of}$ \quad $\underset{Death}{\underline{\mathrm{death}}}$ \\
        Qualifier & Severity & $\underset{Severity[mild]}{\underline{\mathrm{mild}}}$ \quad $\xleftarrow[Severity]{}$ \quad $\underset{Observation}{\underline{\mathrm{symptoms}}}$ \\
         & Stability & $\underset{Observation}{\underline{\mathrm{hemodynamically}}}$ \quad $\xrightarrow[Stability]{}$ \quad $\underset{Stability[change]}{\underline{\mathrm{unstable}}}$ \\[2ex]
    
    \hline
    
     &
        After & $\underset{Condition}{\underline{\mathrm{infected}}}$ \quad $\xrightarrow[After]{}$ \quad $\mathrm{following}$ \quad $\underset{Encounter[inpatient]}{\underline{\mathrm{admission}}}$ \\
        
        & Before & $\mathrm{diagnosis of}$ $\underset{Condition}{\underline{\mathrm{aortic\ stenosis}}}$ \quad $\xrightarrow[Before]{}$ \quad $\mathrm{prior\ to}$ \quad $\underset{Encounter}{\underline{\mathrm{visit}}}$ \\
        
        & Duration & $\underset{Condition}{\underline{\mathrm{type\ 1\ diabetes}}}$ \quad $\xrightarrow[Duration]{}$ \quad $\mathrm{for}$ \quad $\underset{Eq-Comparison}{\underline{\mathrm{at\ least\ 1\ year}}}$ \\ 
        Temporal and Comparative & During & $\underset{Procedure}{\underline{\mathrm{mechanically \ ventilated}}}$ \quad $\xrightarrow[During]{}$ \quad $\mathrm{while}$ \quad $\underset{Encounter[inpatient]}{\underline{\mathrm{admitted}}}$ \\
        
         & Numeric-Filter & $\underset{Observation[vital]}{\underline{\mathrm{body\ weight}}}$ \quad $\xrightarrow[Numeric-Filter]{}$ \quad $\underset{Eq-Comparison}{\underline{\mathrm{less\ than\ 110\ pounds}}}$ \\    
        
         & Minimum-Count & $\underset{Encounter[inpatient]}{\underline{\mathrm{admitted}}}$ \quad $\xrightarrow[Minimum-Count]{}$ \quad $\underset{Eq-Comparison}{\underline{\mathrm{at\ least\ twice}}}$ \\    
        
         & Temporality & $\underset{Encounter}{\underline{\mathrm{seen}}}$ \quad $\xrightarrow[Temporality]{}$ \quad $\underset{Eq-Comparison}{\underline{\mathrm{within\ past\ 6\ months}}}$ \\[2ex]
         
    \hline
    
    Other &
         Location & $\underset{Encounter[inpatient]}{\underline{\mathrm{admitted}}}$ \quad $\xrightarrow[Location]{}$ \quad $\mathrm{to\ the}$ \quad $\underset{Location[unit]}{\underline{\mathrm{ICU}}}$ \\[2ex]
    
\end{tabular}
    \caption{\textbf{Examples of representative relations.} Direction of arrows indicates role, i.e., subject → target entity.}
    \label{tbl_relation_examples}
\end{table*}

\noindent The LCT annotation schema contributes the following novel features: (1) deep granularity in entities and relations, which enables (2) rich semantic representation, closely capturing the intent of complex clinical trial eligibility criteria and facilitating accurate query generation.

\subsubsection*{Deep Entity and Relation Granularity}
\noindent We assume that more specific annotation labels are generally more straightforward to generate accurate queries with. For example, within the span, "preceding six months", annotating the token "preceding" as \textit{Temporal} (an entity type in Chia) may appear to be adequate, given that an English-speaking human would understand that this refers to the past. Without further information, however, a naïve algorithm would be unable to determine (1) whether such a entity refers to the past, present, or future, (2) that the token "six" refers to a numeric value, and (3) that "months" refers to a unit of temporal measurement. In such cases, most query generation algorithms introduce additional rule-based or syntactic parsing modules, such as SuTime \cite{chang2012sutime} to further normalize the phrase to a value \cite{weng2011elixr, yuan2019criteria2query}. This ambiguity in label semantics creates unnecessary complexity in downstream systems, requiring that the same text be processed a second time. \\

\noindent In contrast, we designed the LCT annotation schema to favor discrete, explicit entities and relations where possible, with an aim toward reducing the need for additional normalization steps needed for query generation. In our annotation schema, this example would be annotated with the following fine-grained entities: \\

\begin{center}
\begin{tabular}{c c c}
    "preceding" & "six" & "months" \\ 
    \big\downarrow & \big\downarrow & \big\downarrow \\
    \textit{Eq-Temporal-Period} & \textit{Eq-Value} & \textit{Eq-Temporal-Unit} \\
    \textit{[past]} & & \textit{[month]} \\
\end{tabular}
\end{center}

\noindent As shown in the example, each token is uniquely annotated, with the values \textit{[past]} and \textit{[month]} serving to more clearly disambiguate semantics of the temporal phrase to a normalized temporal value. Moreover, as fine-grained entities are connected by relations to general entities, which can in turn have relations to other general entities, the LCT annotation schema is able to capture eligibility criteria semantics at a deeper level than other corpora. 

\subsubsection*{Rich Semantic Representation}
\noindent Certain eligibility criteria cannot be directly translated into queries, but instead must first be reasoned upon. For example, a query to find patients meeting the criterion of "conditions contraindicating pioglitazone" requires reasoning to first answer the question, \textit{What} conditions contraindicate use of pioglitazone? Such reasoning may be performed by a knowledge base or other methods, but cannot be done unless the contraindicative relation is detected:

\begin{center}
\begin{tabular}{c c c c c}
    "Conditions" & & "contraindicating" & & "pioglitazone" \\ 
    \big\downarrow & & \big\downarrow & & \big\downarrow \\
    \textit{Condition} & $\xleftarrow[Caused-By]{}$ & \textit{Contraindication} & $\xrightarrow[Contraindicates]{}$ & \textit{Drug} \\[-1ex]
    & & & & + \\
    & & & & \textit{Drug-Name} \\
\end{tabular}
\end{center}

\noindent As the span "Conditions" is labeled \textit{Condition} but does not have an overlapping \textit{Condition-Name} entity, it is considered unnamed and thus would need to be reasoned upon to determine. "[P]ioglitazone", on the other hand, includes a \textit{Drug-Name} entity and is thus considered named. The absence of overlapping \textit{Name} entities serves as an indicator to downstream applications that reasoning may be needed to determine relevant conditions or drugs. We examine additional cases of the LCT's semantic representation and benefits in the next section, where we compare the LCT annotation schema to Chia's.

\begin{figure}[ht!]
  \includegraphics[scale=0.56]{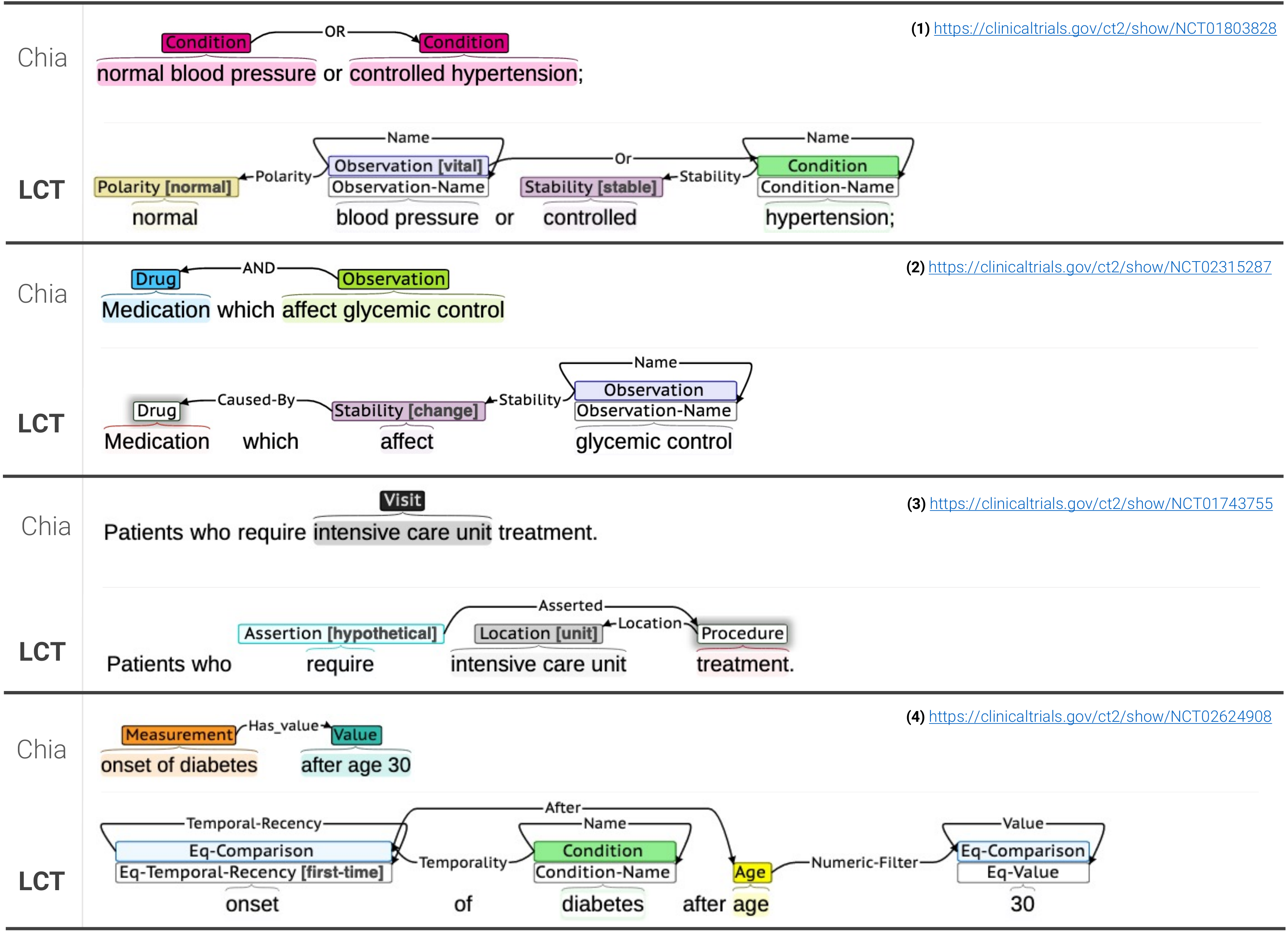}  
    \caption{Examples of clinical trials eligibility criteria annotated with Chia and LCT annotation schemas. Each example shows a criterion from a Chia annotation (above) and an LCT annotation of the same text for purposes of comparison (below).}
    \label{fig_chia_vs_lct}
\end{figure}

\subsection*{Comparison to Chia}
We designed the LCT annotation schema by building upon the important previous work of EliIE and Chia. As Chia itself builds upon EliIE and is more recent, in the next section we compare the LCT corpus and Chia by examining cases of ambiguity handling and annotation difference in entities, relations, and coupling of entity types to the OMOP \cite{hripcsak2015observational} (Observational Medical Outcomes Partnership) data model. For each case we examine the LCT annotation schema's novel solutions and contributions. Figure \ref{fig_chia_vs_lct} shows comparison examples of annotations of the same eligibility criteria using the two corpora in the Brat annotation tool \cite{stenetorp2012brat}.

\subsubsection*{Capturing Entity Semantics}
\noindent Example 2 of Figure \ref{fig_chia_vs_lct} demonstrates the need to closely capture semantics in clinical trials eligibility criteria for unnamed entities. The span "Medication" in "Medication[s] which affect glycemic control" refers to \textit{any} drug which potentially affects glycemic control. As discussed, the LCT annotation schema uses \textit{Name} entities to handle such cases, where the absence in this example of a \textit{Drug-Name} entity indicates that "Medication" refers to any drug, and thus may need to be determined by downstream use of a knowledge base or other methods. \\

\noindent As can be seen, Chia does not differentiate between named and unspecified drugs, conditions, procedures and so on. While it is true that for query generation one may need to normalize these spans to coded representations (e.g., ICD-10, RxNorm or LOINC codes) and may in the process find that the span "Medication" is not a particular medication (and thus can be assumed to be \textit{any} medication), such a workaround nonetheless complicates usage of the corpus in finding and handling such cases in a more direct, less error-prone way. \\

\noindent Consider also the phrase, "after age 30" in example 4 in Figure \ref{fig_chia_vs_lct}. In Chia, disambiguation of time units, values, and chronological tense must be performed by additional processing, as the Chia \textit{Value} entity provides no information as to the semantics of the component sub-spans. In contrast, in the LCT corpus the tokens "after", "age", "30" are annotated with the explicit entities and relations to enable more straightforward query generation. \\

\subsubsection*{Capturing Relation Semantics}
Eligibility criteria frequently contain entities which relate to other entities in the form of examples, abbreviations, equivalencies, or explicit lists. Chia uses \textit{Subsumes} relations to denote that one or more entities are a subset, depend on, or are affected by another entity in some way. In many cases however these entities leave significant semantic ambiguity which may complicate query generation. Consider the phrase:

\begin{quote} 
\centering 
\textit{"conditions predisposing to dysphagia (eg, gastroesophageal reflux disease [GERD] ...)"}
\end{quote} 

In this case, "gastroesophageal reflux disease" is an \textit{example} of a condition, while "GERD" is an \textit{abbreviation}. However, both are \textit{Subsumes} relations in the Chia annotation:

\begin{center}
\begin{tabular}{c c c c c c c c c c c c}
    \multicolumn{5}{c}{$\underbrace{\text{"conditions predisposing to dysphagia"}}$} & "(eg," & \multicolumn{3}{c}{$\underbrace{\text{"gastroesophageal reflux disease"}}$} & & "[GERD]" \\
    \multicolumn{5}{c}{\big\downarrow}                                            &      & \multicolumn{3}{c}{\big\downarrow} & & \big\downarrow & \\
    \multicolumn{5}{c}{\textit{Condition}}                                        & $\xrightarrow[Subsumes]{}$ & \multicolumn{3}{c}{\textit{Condition}} & $\xrightarrow[Subsumes]{}$ & \textit{Condition} \\
\end{tabular}
\end{center}

\noindent In contrast, in the LCT annotation schema this example would be annotated as:

\begin{center}
\begin{tabular}{c c c c c c c c c}
    \text{"conditions"} & \text{"predisposing"} & "to" & \text{"dysphagia"} & 
    "(eg," & \multicolumn{3}{c}{$\underbrace{\text{"gastroesophageal reflux disease"}}$} & \text{"[GERD]"} \\
    
    \big\downarrow & \big\downarrow & & \big\downarrow & & \multicolumn{3}{c}{\big\downarrow} & \big\downarrow \\
    
    \textit{Condition} & \makecell[t]{\textit{Assertion} \\ \textit{[hypothetical]}} & $\xrightarrow[Asserted]{}$ & \textit{Condition} & &
    \multicolumn{3}{r}{\textit{Condition}} $\xleftarrow[Abbrev-Of]{}$ & \textit{Condition} \\ [-3ex]
    
     & & & + & & \multicolumn{3}{c}{+} & + \\
    
    & & & \textit{Condition-Name} & & \multicolumn{3}{c}{\textit{Condition-Name}} & \textit{Condition-Name} \\
\end{tabular}
\end{center}

\begin{tikzpicture}
    [
      remember picture,
      overlay,
      -latex,
      color=black!75,
      yshift=-5.2ex,
      xshift=9.8ex,
      shorten >=1pt,
      shorten <=1pt,
    ]
    \draw[thick,->] (5.55,1) -| (5.55,0.8) -- (-1,0.8) node [below,pos=0.5] {\footnotesize{\textit{Caused-By}}} -| (-1,1.9);
    \draw[thick,->] (10.53,1) -| (10.53,0.2) -- (-1,0.2) node [below,pos=0.5] {\footnotesize{\textit{Example-Of}}} -| (-1,0.8);
  \end{tikzpicture} \\ \\ \\

\noindent The LCT annotation uses \textit{Abbrev-Of} and \textit{Example-Of} relations to clearly differentiate relations between "gastroesophageal reflux disease", "GERD", and "conditions predisposing to dysphagia". Additionally, rather than grouping the latter into a single \textit{Condition} entity, the LCT is also much more granular, with the annotation reflecting that dysphagia is a condition patients are hypothetically predisposed to (due to other conditions such as GERD), but not necessarily actively afflicted by. \\

\noindent Another example illustrating the importance of capturing relation semantics can be seen in the following Chia annotations: \\

\begin{center}
\begin{tabular}{l c c c}
    (1) & $\underbrace{\text{"type 1 diabetes"}}$ & & $\underbrace{\text{"for at least 1 year"}}$ \\ 
    & \big\downarrow & & \big\downarrow \\
    & \textit{Condition} & $\xrightarrow[Has-Temporal]{}$ & \textit{Temporal}
\end{tabular}
\end{center}

\begin{center}
\begin{tabular}{l c c c}
    (2) & $\underbrace{\text{"Acute coronary syndrome"}}$ & & $\underbrace{\text{"in the past 6 months"}}$ \\ 
    & \big\downarrow & & \big\downarrow \\
    & \textit{Condition} & $\xrightarrow[Has-Temporal]{}$ & \textit{Temporal}
\end{tabular}
\end{center}

\noindent While syntactically similar, the semantics in chronology expressed in the two criteria are different. In (1), "type 1 diabetes for at least 1 year" suggests that the diagnosis of type 1 diabetes mellitus should have occurred at least 1 year prior to the present. In other words, a unit of temporal measurement (1 year), should have passed since initial diagnosis. In contrast, "Acute coronary syndrome in the past 6 months" (2) suggests that a range of dates between the present and a past event (past 6 months), should have passed since the diagnosis. In Chia, however, the same \textit{Has-Temporal} relation is used for both, blurring distinctions between \textit{durations of time} versus \textit{ranges of dates}, potentially leading to errors during query generation.  \\

\noindent In the LCT annotation schema, these would be annotated as (omitting fine-grained entities for brevity): \\

\begin{center}
\begin{tabular}{l c c c}
    (1) & $\underbrace{\text{"type 1 diabetes"}}$ & "for" & $\underbrace{\text{"at least 1 year"}}$ \\ 
    & \big\downarrow & & \big\downarrow \\
    & \textit{Condition} & $\xrightarrow[Duration]{}$ & \textit{Eq-Comparison} \\[-1ex]
    & + & & \\
    & \textit{Condition-Name} & &
\end{tabular}
\end{center}

\begin{center}
\begin{tabular}{l c c c}
    (2) & $\underbrace{\text{"Acute coronary syndrome"}}$ & "in the" & $\underbrace{\text{"past 6 months"}}$ \\ 
    & \big\downarrow & & \big\downarrow \\
    & \textit{Condition} & $\xrightarrow[Temporality]{}$ & \textit{Eq-Comparison} \\[-1ex]
    & + & & \\
    & \textit{Condition-Name} & &
\end{tabular}
\end{center}

\noindent The LCT annotations distinguish these types of temporal semantics by using distinct \textit{Duration} and \textit{Temporality} relations, allowing downstream queries to more accurately reflect researcher intent. The LCT corpus also does not include "for" or "in the" as part of the entities.

\subsubsection*{Data Model Mapping}
The Chia annotation schema is mapped to the OMOP Common Data Model \cite{hripcsak2015observational} and is designed to ease integration with other OMOP-related tools and generation of SQL queries on OMOP databases. Chia OMOP-derived entities generally follow the naming convention of OMOP domains and SQL database tables, such as \textit{Person}, \textit{Condition}, \textit{Device} and so on. \\

\noindent The LCT annotation schema takes a different approach by intentionally avoiding direct mappings to data models. This approach was chosen to (1) allow the annotation entities and relations flexibility to be transformed to any data model (including but not limited to OMOP) and (2) provide flexibility in capturing criteria important to the task of query generation, even when such criteria are not represented in OMOP. \\

\noindent A disadvantage of directly coupling an annotation schema to a data model is evidenced by criteria such as:

\begin{quote} 
\centering 
\textit{"Males aged 18 years and above"}
\end{quote}

\noindent In Chia, spans related to gender and age share the same \textit{Person} entity: \\

\begin{center}
\begin{tabular}{c c c c c c c}
    "Males" & "aged" & & \multicolumn{4}{c}{$\underbrace{\text{"18 years or above"}}$} \\ 
    \big\downarrow & \big\downarrow & & \multicolumn{4}{c}{\big\downarrow}  \\
    \textit{Person} &\textit{Person} & $\xrightarrow[Has\_value]{}$ & \multicolumn{4}{c}{\textit{Value}} \\
\end{tabular}
\end{center}

\noindent The use of the Chia \textit{Person} entity across gender and age results in loss of information and complications for query generation. As with quantitative and temporal annotations, the generic \textit{Person} entity again forces the burden of normalization and additional parsing to downstream applications. In contrast, this example would be annotated first using general entities and relations in the LCT annotation schema:

\begin{center}
\begin{tabular}{c c c c c c c c}
    "Males" & "aged" & & \multicolumn{4}{c}{$\underbrace{\text{"18 years or above"}}$} \\ 
    \big\downarrow & \big\downarrow & & \multicolumn{4}{c}{\big\downarrow}  \\
    \textit{Life-Stage-And-Gender[male]} &\textit{Age} & $\xrightarrow[Numeric-Filter]{}$ & \multicolumn{4}{c}{\textit{Eq-Comparison}} \\
\end{tabular}
\end{center}

\noindent Followed by fine-grained entities and values:

\begin{center}
\begin{tabular}{llll}
    \multirow{4}{5.5em}[-3pt]{\textit{\mbox{Eq-Comparison}}} & \multirow{4}{1em}[0pt]{$\begin{cases}\\\\\\\\\end{cases}$} & $\xrightarrow[Value]{}$ & \textit{Eq-Value "18"} \\
    & & $\xrightarrow[Temporal-Unit]{}$ & \textit{Eq-Temporal-Unit[year]} "years" \\
    & & $\xrightarrow[Operator]{}$ & \textit{Eq-Operator[GTEQ]} "or above" \\
\end{tabular}
\end{center}

\noindent The LCT annotation captures the male and age spans as distinguishable entities, closely preserving the semantics of the original text. 

\subsection*{Annotation process}
We used eligibility criteria from \url{https://clinicaltrials.gov} as the basis for our corpus. \\

\noindent We extracted 1,020 randomly selected clinical trials eligibility descriptions, 20 for training and inter-annotator comparison and 1,000 for post-training annotation. Documents were included only if they met the following criteria:
\begin{enumerate}
    \item The combined inclusion and exclusion criteria text was at least 50 characters long.
    \item The clinical trial was uploaded on or after January 1st, 2018. This date was chosen because we found that clinical trials performed further in the past appeared to exhibit less structural consistency in language, punctuation and indentation compared to more recent text.
\end{enumerate}

\noindent During annotation, 14 documents were found to be information poor (often with no spans to annotate) and discarded, resulting in 1,006 total annotated eligibility descriptions. Annotation was performed by two annotators, the first a biomedical informatician and the second a computer scientist. For initial annotation training, 20 documents were distributed to both annotators. Annotation was done in the following steps:

\begin{enumerate}
    \item Annotation meetings were held bi-weekly for 3 months following initial annotation training in which the annotation guidelines were introduced. Initial meetings focused on discussion of annotation guideline implementation and revision. After each meeting, the annotation guidelines were revised to include new named entities and relationships and inter-annotator agreement was recalculated using F\textsubscript{1}-scores. Each annotator used the UMLS Terminology Services (UTS) Metathesaurus Browser (\url{https://https://uts.nlm.nih.gov/uts/umls/home}) to search for biomedical concepts whose meaning was unclear.
    \item After annotation guideline revisions and annotation training were completed, eligibility criteria were assigned to each annotator, with each clinical trial eligibility criteria annotated by a single annotator using the BRAT annotation tool \cite{stenetorp2012brat}. Due to differences in time availability for annotation, roughly 90\% (887 documents) of the annotation task was performed by the first annotator, and 99 documents by the second annotator.
    \item At the point in which 50\% of the corpus was annotated, we trained two neural networks (one for general entities and another for fine-grained entities) using the NeuroNER tool \cite{dernoncourt2017neuroner} on our manually annotated eligibility criteria to predict annotations for the remaining 50\%. NeuroNER is a state-of-the-art entity extraction system which has been successfully adapted to tasks such as de-identification and concept extraction. NeuroNER utilizes bidirectional Long Short-Term Memory and Conditional Random Fields (biLSTM+CRF) for token-level multiple-label prediction. We used the NeuroNER-predicted entities to auto-populate our remaining eligibility descriptions.
    \item Manual annotation was completed on the remaining 50\% of eligibility descriptions by editing and correcting the predicted entities from NeuroNER in (3).

\end{enumerate}    
    
\noindent The resulting corpus included 887 single-annotated and 119 double-annotated total notes. 

\section*{Data Records}
\label{sec:data_records}
The LCT corpus annotated eligibility criteria and text documents can be found at \url{https://doi.org/10.6084/m9.figshare.17209610}\cite{dobbins2022_lct_figshare}. Code for pre-annotation and analysis are available at \url{https://github.com/uw-bionlp/clinical-trials-gov-data}. The LCT corpus is annotated using the Brat "standoff" format. The Brat format includes two file types, ".txt" files and ".ann" files. \\

\subsection*{Text (.txt) files}
The free-text eligibility criteria information in the 1,006 documents of the LCT corpus. Each file is named using the "NCT" identifier used by \url{https://clinicaltrials.gov}. \\

\subsection*{Annotation (.ann) files}
The annotation files used by Brat for tracking annotated spans of text and relations. Each .ann file corresponds to a .txt file of the same name. Each row of a .ann file may begin with a "T" (for an entity) or "R" (for a relation), followed by an incremental number for uniquely identifying the entity or relation (e.g., "T15"). "T" rows are of the form "T<number> <entity type> <start character index> <stop character index>", where start and stop indices correspond to text in the associated .txt file. "R" rows are of the form "R<number> <relation type> Arg1:<ID> Arg2:<ID>", where ID values correspond to identifiers of entities. Additionally, for ease of annotation certain LCT relations are defined as arguments of Brat "events", identified by "E". "E" rows are of the form "E<number> <entity type>:<ID> <relation type>:<ID>". \\

\noindent More information on the Brat format can be found at \url{https://brat.nlplab.org/standoff.html}. \\

\section*{Technical Validation}
\label{sec:technical_validation}
\subsection*{Inter-annotator agreement}
\noindent Inter-annotator agreement was calculated using F\textsubscript{1} scoring for entities and relations with 20 double-annotated documents. Entity annotations were considered matching only if entity types and token start and end indices matched exactly. Relations annotations were similarly considered matching only if relation type and token start and end indices of both the subject and target matched exactly. \\

\noindent Initial inter-annotator agreement using the 20 training documents was 76.1\% for entities and 60.3\% for relations. Inter-annotator agreement improved slightly to 78.1\% (+2\%) for entities and 60.9\% (+0.6\%) for relations in the 99 additional double-annotated documents, indicating reasonably high annotator agreement considering the complexity of the annotation task. \\

\noindent We found two categories of annotation differences in double-annotated eligibility criteria. First, in spans such as:

\begin{quote} 
\centering 
\textit{"Reported being a daily smoker"}
\end{quote}

\noindent While both annotators tended to annotate "smoker" as both a \textit{Condition} and \textit{Condition-Name}, adjectives such as "daily" were often annotated as \textit{Eq-Comparison} and \textit{Eq-Temporal-Unit[day]} by one annotator and \textit{Stability[stable]} by the other. After review, these were generally reconciled to the first pattern. \\

\noindent The second category of differences can be seen in spans such as "CDI Diarrhea", where "CDI" refers to Clostridium Difficile Infection. In these cases, the annotators may annotate this span as (omitting \textit{Condition-Name} entities for brevity):

\begin{center}
\begin{tabular}{l c c c}
    (1) & "CDI" & & "Diarrhea" \\ 
    & \big\downarrow & & \big\downarrow \\
    & \textit{Condition} & $\xleftarrow[Caused-By]{}$ & \textit{Condition} \\
\end{tabular}
\end{center}

\begin{center}
    or
\end{center}

\begin{center}
\begin{tabular}{l c}
    (2) & "CDI Diarrhea" \\ 
    & \big\downarrow \\
    & \textit{Condition} \\
\end{tabular}
\end{center}

\noindent The first annotation separates "CDI" and "Diarrhea" into two entities, with "Diarrhea" \textit{Caused-By} "CDI", while the second annotation treats them as a single entity. Reconciliation in these cases was done by referring to the UMLS Metathesaurus to determine whether the combined span existed as a single concept within the UMLS. As "Clostridium Difficile Diarrhea" exists as a UMLS concept (C0235952), the annotations in this example were reconciled to use the second, multi-span entity. 

\subsection*{Baseline prediction}
\noindent To evaluate baseline predictive performance on the LCT corpus, we first created a randomly assigned 80/20 split of the corpus, with 804 documents used for the training set and 202 for the test set. For entity prediction, we trained NER models using biLSTM+CRF and BERT \cite{devlin2018bert} neural architectures. For BERT-based prediction, we used two pretrained models trained on published medical texts, SciBERT \cite{beltagy2019scibert} and PubMedBERT \cite{gu2021domain}. For both biLSTM+CRF and BERT predictions, we trained one model to predict general entities and another for fine-grained entities. \\

\noindent For relation extraction, we evaluated SciBERT for sequence classification as well as a modified BERT architecture, R-BERT, following methods developed by Wu \& He \cite{wu2019enriching}, also using the pretrained SciBERT model. Table \ref{tbl_hyperparams} shows hyperparameters used for each task. \\

\def\arraystretch{1.2}
\begin{table}[h!]
\begin{tabular}{m{4cm} m{3cm} m{5cm} m{3cm}}
 \toprule
 \textbf{Task} & \textbf{Architecture} & \textbf{Hyperparameter / Embeddings} & \textbf{Training Value} \\
 \hline
    \multirow{4}{*}{\mbox{Named Entity Recognition}} &
    \multirow{4}{*}{\mbox{biLSTM+CRF}} 
        & Character Dimensions & 25 \\
        & & Token Embedding Dimensions & 100 \\
        & & Learning Rate & 0.005 \\
        & & Dropout & 0.5 \\
        & & Pretrained Embeddings & GloVe \cite{pennington2014glove} \\
    \hline
    \multirow{2}{*}{\mbox{Relation Extraction}} &
    \multirow{2}{*}{\mbox{BERT \& R-BERT}} 
        & Pretrained Model & SciBert  \\
        & & Learning Rate & 0.00003 \\
 \hline
\end{tabular}
\caption{Hyperparameters and pre-trained embeddings used for named entity recognition and relation extraction baseline results. For the NER task, the same architecture and hyperparameters were used for both general and fine-grained entity models. For the relation extraction task, the same hyperparameters were used with both the BERT and R-BERT architectures.}
\label{tbl_hyperparams}
\end{table}

\noindent  We achieved the highest micro-averaged F\textsubscript{1} score of 81.3\% on entities using SciBERT and 85.2\% on relations using the R-BERT architecture with SciBERT. Results of representative entities and relations are shown in Tables \ref{entity_f1} and \ref{relation_f1}. \\

\begin{table}[tp]
    \def\arraystretch{1.4}
\begin{tabular}{m{2cm} m{3.2cm} m{1.4cm} m{2.7cm} m{2.7cm} m{2.7cm}}
    \toprule
    \textbf{Category} & \textbf{Entity} & \textbf{Count} & \textbf{biLSTM+CRF} & \textbf{PubMedBERT} & \textbf{SciBERT} \\ \midrule
     & Condition & 7,087 & 78.6 / 78.1 / 78.3 & 76.1 / 79.4 / 77.7 & 78.4 / 83.3 / 80.8 \\
     & Contraindication & 142 & 93.7 / 78.9 / 85.7 & 77.4 / 80.0 / 78.6 & 100.0 / 96.6 / 98.3 \\
    Clinical & Drug & 1,404 & 76.8 / 81.3 / 79.0 & 74.1 / 80.9 / 77.4 & 73.4 / 80.9 / 77.0 \\
     & Encounter & 302 & 64.1 / 58.1 / 60.9 & 51.7 / 61.7 / 56.3 & 58.3 / 74.4 / 65.4 \\
     & Observation & 2,558 & 74.3 / 66.1 / 69.9 & 67.9 / 73.5 / 70.6 & 72.1 / 77.6 / 74.7 \\
     & Procedure & 3,016 & 68.4 / 75.5 / 71.9 & 67.0 / 75.9 / 71.2 & 71.3 / 79.4 / 75.1 \\
    \hline       
    \multirow{4}{*}[-4pt]{\mbox{Demographic}} & 
        Age & 708 & 91.3 / 95.4 / 93.3 & 82.4 / 88.5 / 85.3 & 99.1 / 98.3 / 98.7 \\
     & Birth & 27 & 100.0 / 80.0 / 88.8 & 100.0 / 62.5 / 76.9 & 100.0 / 62.5 / 76.9 \\
     & Death & 35 & 33.3 / 33.3 / 33.3 & 0.0 / 0.0 / 0.0 & 100.0 / 20.0 / 33.3 \\
     & Family-Member & 147 & 40.0 / 19.0 / 25.8 & 33.3 / 55.5 / 41.6 & 44.9 / 61.1 / 51.7 \\
     & Language & 194 & 92.5 / 96.1 / 94.3 & 73.8 / 100.0 84.9 & 96.6 / 93.5 / 95.0 \\
    \hline
    Logical & Negation & 952 & 74.3 / 82.7 / 78.2 & 60.9 / 73.1 / 66.4 & 73.5 / 82.9 / 77.9 \\
    \hline
    \multirow{6}{*}[-5pt]{\mbox{Qualifier}} &
        Assertion & 1,157 & 66.6 / 62.8 / 64.7 & 56.1 / 58.9 / 57.5 & 62.1 / 65.8 / 63.9 \\
         & Modifier & 3,464 & 65.0 / 58.3 / 61.5 & 59.2 / 64.0 / 61.5 & 58.5 / 65.4 / 61.8 \\
         & Polarity & 360 & 82.5 / 88.0 / 85.1 & 74.6 / 67.4 / 70.8 & 81.4 / 79.5 / 80.4 \\
         & Risk & 117 & 93.1 / 96.4 / 94.7 & 91.3 / 91.3 / 91.3 & 95.4 / 91.3 / 93.3 \\
         & Severity & 569 & 86.8 / 90.8 / 88.7 & 76.7 / 79.5 / 78.1 & 86.5 / 94.1 / 90.2 \\
         & Stability & 397 & 84.2 / 67.6 / 75.0 & 79.4 / 75.0 / 77.1 & 75.3 / 84.7 / 79.7 \\
    \hline
      &
        Criteria-Count & 33 & 50.0 / 66.6 / 57.1 & 28.5 / 40.0 / 33.3 & 12.5 / 20.0 / 15.5 \\
     & Eq-Comparison & 5,298 & 83.1 / 83.8 / 83.4 & 81.4 / 85.0 / 83.2 & 85.3 / 89.3 / 87.3 \\
     Temporal and & Eq-Temporal-Period & 2,057 & 88.7 / 89.2 / 88.9 & 70.0 / 73.9 / 71.9 & 82.6 / 86.3 / 84.4 \\
     Comparative & Eq-Temporal-Recency & 131 & 68.7 / 84.6 / 75.8 & 43.4 / 55.5 / 48.7 & 50.0 / 66.6 / 57.1 \\
     & Eq-Temporal-Unit & 1,808 & 95.1 / 97.6 / 96.4 & 97.4 / 98.1 / 97.8 & 98.2 / 99.4 / 98.8 \\
     & Eq-Value & 3,835 & 91.8 / 95.3 / 93.5 & 95.5 / 96.2 / 95.9 & 96.4 / 97.1 / 96.7  \\
    \hline   
    Other &
        Location & 371 & 68.5 / 58.7 / 63.2 & 65.4 / 71.6 / 68.3 & 73.4 / 78.3 / 75.8 \\
    \hline
    - & Total & 56,146 & 80.2 / 79.6 / 79.9 & 75.3 / 78.7 / 77.0 & 79.0 / 83.7 / 81.3 \\
    
\end{tabular}

    \caption{\textbf{Baseline entity prediction scores (\%, Precision / Recall / F\textsubscript{1}).} Corpus-level micro-averaged scores are shown in the bottom row. For brevity a representative sample of entities is shown. \textit{Count} refers to the total count of unique spans annotated in the entire corpus. Entities included in the total count and scores but omitted for brevity are \textit{Acuteness, Allergy, Condition-Type, Code, Coreference, Ethnicity, Eq-Operator, Eq-Unit, Indication, Immunization, Insurance, Life-Stage-And-Gender, Organism, Other, Specimen, Study and Provider}.}
    \label{entity_f1}
\end{table}

\begin{table*}
    \centering
    \def\arraystretch{1.4}
\begin{tabular}{m{4.5cm} m{2.5cm} m{2cm} m{2.8cm} m{3cm}}
\toprule
    \textbf{Category} & \textbf{Relation} & \textbf{Count} & \textbf{SciBERT} & \textbf{R-BERT+SciBERT} \\ \midrule
     & Abbrev-Of &           462 & 95.2 / 90.9 / 93.0 & 92.3 / 93.1 / 94.2 \\
    Alternatives and Examples & Equivalent-To & 516 & 61.5 / 69.5 / 65.3 & 59.6 / 67.3 / 63.2 \\
     &                       Example-Of & 1,497 & 94.8 / 92.9 / 93.8 & 90.5 / 91.7 / 91.1 \\
    \hline
     &                       Contraindicates & 153 & 90.9 / 90.9 / 90.9 & 90.9 / 90.9 / 90.9 \\
     &                       Caused-By & 726 & 63.0 / 86.4 / 72.9 & 78.6 / 86.4 / 82.3 \\
     Clinical &              Found-By & 293 & 90.4 / 59.3 / 71.7 & 79.3 / 71.8 / 75.4 \\
     &                       Treatment-For & 457 & 69.2 / 69.2 / 69.2 & 61.7 / 74.3 / 67.4 \\
     &                       Using & 405 & 73.8 / 83.7 / 78.4 & 66.6 / 64.8 / 65.7 \\
    \hline
    \multirow{4}{*}[0pt]{\mbox{Logical}} & And & 821 & 54.1 / 60.0 / 56.9 & 53.8 / 53.8 / 53.8 \\
     &                       If-Then & 261 & 57.6 / 65.2 / 61.2 & 55.5 / 65.2 / 60.0 \\
     &                       Negates & 984 & 74.3 / 91.0 / 81.8 & 74.5 / 88.7 / 81.0 \\
     &                       Or & 4,156 & 85.1 \ 93.2 \ 89.0 & 88.4 / 92.2 / 90.2 \\
    \hline
     &               Asserted & 1,184 & 83.7 / 89.0 / 86.3 & 85.9 / 89.0 / 87.5 \\
     &                       Modifies & 3,400 & 90.9 / 94.2 / 92.5 & 92.2 / 95.4 / 93.8 \\
    Qualifier &               Risk-For & 90 & 92.3 / 85.7 / 88.8 & 92.8 / 92.8 / 92.8 \\
     &                       Severity & 529 & 80.2 / 96.6 / 87.6 & 86.3 / 96.6 / 91.2 \\
     &                       Stability & 395 & 76.0 / 92.6 / 83.5 & 76.4 / 95.1 / 84.7 \\
    \hline
    \multirow{6}{*}[-5pt]{\mbox{Temporal and Comparative}} & After & 166 & 75.0 / 70.5 / 72.7 & 72.2 / 76.4 / 74.2 \\
     &                       Before & 320 & 70.2 / 86.6 / 77.6 & 78.1 / 83.3 / 80.6 \\
     &                       Duration & 243 & 59.3 / 79.1 / 67.8 & 64.5 / 83.3 / 72.7 \\
     &                       During & 350 & 66.6 / 68.7 / 67.6 & 63.6 / 65.6 / 64.6 \\
     &                       Numeric-Filter & 1,957 & 84.6 / 93.3 / 88.7 & 85.7 / 92.3 / 88.8 \\
     &                       Minimum-Count & 173 & 64.2 / 69.2 / 66.7 & 71.4 / 76.9 / 74.0 \\
     &                       Temporality & 2,645 & 80.7 / 90.7 / 85.4 & 81.8 / 92.2 / 86.7 \\
    \hline
    Other &                  Location & 207 & 64.2 / 94.7 / 76.6 & 69.2 / 94.7 / 80.0 \\
    \hline
    - & Total & 24,379 & 80.2 / 88.2 / 84.0 & 82.5 / 88.0 / 85.2
\end{tabular}

    \caption{\textbf{Baseline relation prediction scores (\%, Precision / Recall / F\textsubscript{1}).} Corpus-level micro-averaged scores are shown in the bottom row. For brevity a representative sample of relations is shown. \textit{Count} refers to the total count annotated in the entire corpus, including relations not shown. The count total excludes general to fine-grained entity relations, which as overlapping spans are not used for relation prediction. Relations included in the total count and scores but omitted for brevity are \textit{Acuteness, Code, Criteria, Except, From, Indication-For, Is-Other, Max-Value, Min-Value, Polarity, Provider, Refers-To, Specimen, Stage, Study-Of and Type}.}
    \label{relation_f1}
\end{table*}

\noindent Among entities, we found two particular categories performed relatively well with F\textsubscript{1} scores of 70\% or often greater: (1) Entities which are syntactically varied but occurred relatively frequently in eligibility descriptions, such as \textit{Condition}, \textit{Procedure}, and \textit{Eq-Comparison}, (2) Entities which sometimes occurred less frequently but with greater relative syntactic consistency and structure, such as \textit{Age}, \textit{Contraindication}, and \textit{Birth}. Entities which occurred very infrequently, such as \textit{Death} tended to have both low precision and recall. \\

\noindent For relations, we found the most frequently occurring relations, such as \textit{Eq-Comparison}, \textit{Temporality}, \textit{Modifies}, and \textit{Example-Of} to perform well, with F\textsubscript{1} scores greater than 85\%. Among less frequently occurring relations, we found a number of cases where relations which tend to occur in similar positions within sentences and grammatical structures were frequently mistaken during prediction. For example, \textit{Eq-Comparison} (e.g., "greater than 40") and \textit{Minimum-Count} (e.g., "at least twice") were sometimes incorrectly predicted. We found similar incorrect predictions for relations such as \textit{Treatment-For} (e.g., "Surgery for malignant pathology") and \textit{Using} (e.g., "knee joint replacement with general anaesthesia"). \\

\noindent In future work we intend to examine approaches improving prediction of less frequently occurring entities and relations. A full listing of baseline prediction results can be found with the annotation guidelines at \url{https://github.com/uw-bionlp/clinical-trials-gov-annotation/wiki/Named-Entity-Recognition-and-Relation-Extraction-performance}. \\

\subsection*{Annotation quality evaluation}
\noindent To determine the quality of single-annotated documents compared to those which were double-annotated, we trained NER models (one for general and another for fine-grained entities, as in earlier experiments) using SciBERT with the 887 single-annotated documents and evaluated on the 119 double-annotated documents. The results were a precision of 79.7\%, recall of 82.5\%, and an F\textsubscript{1} score of 81.4\%, which are very close to the highest performance of our randomly split train/test set results shown in Table \ref{entity_f1}. These results indicate relative uniformity and consistency in the corpus across both single- and double-annotated documents. \\

\def\arraystretch{1.2}
\begin{table}[h!]
\centering
\begin{tabular}{l l c c c}
 \toprule
 \textbf{Training Set} & \textbf{Test Set} & \textbf{Precision} & \textbf{Recall} & \textbf{F\textsubscript{1}} \\
 \hline
    Manual & Semi-automated & 75.4 & 82.1 & 78.6 \\
    Semi-automated & Manual & 80.1 & 79.9 & 80.0 \\
 \hline
\end{tabular}
\caption{\textbf{Results of NER experiments using the manually annotated and semi-automated portions of the corpus.} The manually annotated portion includes 513 documents while the semi-automatically annotated portion is 493 documents.}
\label{tbl_manual_semiauto}
\end{table}

\noindent As the latter near-half (493 documents) of the LCT corpus was automatically annotated, then manually corrected, we also evaluated the quality of the manually annotated portion versus the semi-automatically annotated portion to ensure consistency. We first trained NER models with SciBERT using the manually annotated portion and tested on the semi-automated portion, then reversed the experiment and trained on the semi-automated portion and tested on the manually annotated portion. Results are shown in Table \ref{tbl_manual_semiauto}. \\

\noindent Results of the experiments when training on both the manually and semi-automatically annotated halves of the corpus show comparable results, with the greatest difference being in precision, with the manual annotation-trained model performing slightly worse (-4.7\%) in prediction versus the semi-automated annotation-trained model. Overall F\textsubscript{1} scores were similar at 78.6\% and 80.0\%, suggesting reasonable consistency across the corpus.

\section*{Usage Notes}
\label{sec:usage_notes}
\begin{figure}[p]
\centering
  \includegraphics[scale=0.6]{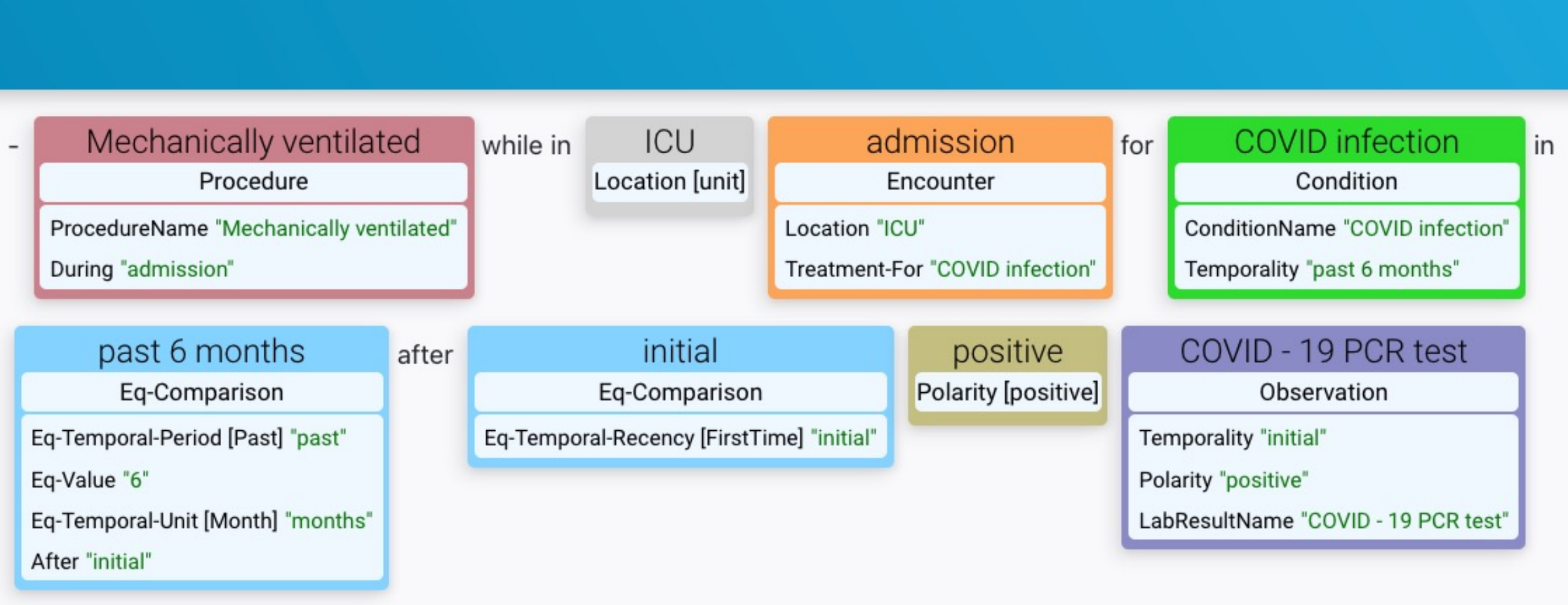}  
\caption{Screenshot of a prototype web application for real-time entity and relation prediction on custom user input text.}
\label{fig_leafai}
\end{figure}

\begin{figure}[p]
\centering
  \includegraphics[scale=0.64]{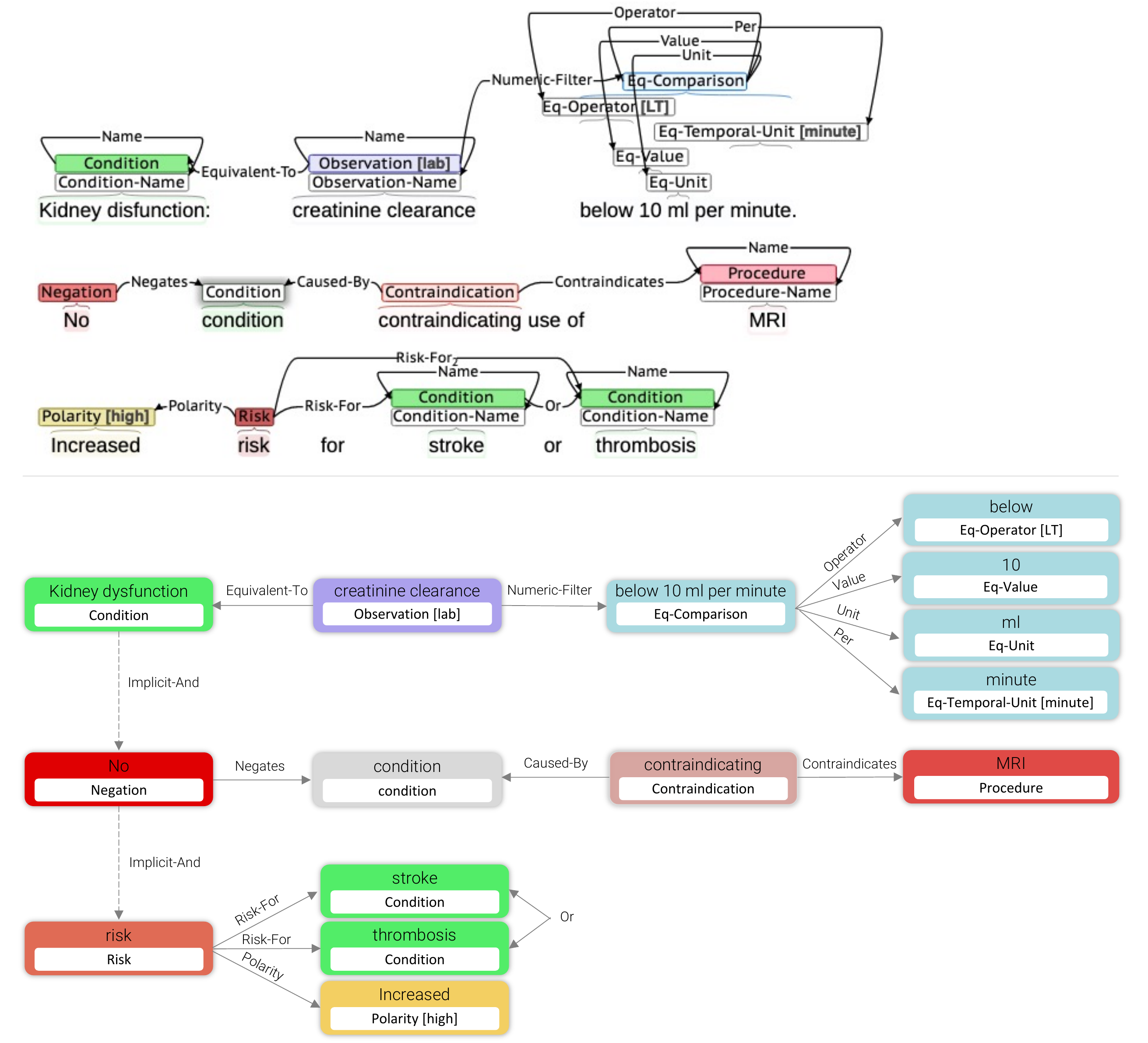}  
\caption{Example of an LCT annotated document (top) transformed into a Directed Acyclical Graph (bottom). LCT entities and relations are readily transformable into tree, graph, or object-oriented representations used for query generation.}
\label{fig_dag}
\end{figure}

The LCT corpus is designed to facilitate query generation and question answering for real-world clinical trials and clinical research, specifically for a future version of the Leaf cohort discovery tool\cite{dobbins2019leaf}. Figure \ref{fig_dag} visualizes an example of a transformation of LCT annotated data into a Directed Acyclical Graph (DAG) structure, which can then be potentially compiled into SQL, FHIR, SPARQL, or other query methods. \\

\noindent To demonstrate the value and utility of the corpus, using the trained baseline Named Entity Recognition and Relation Extraction models, we developed a simple prototype web application to test named entity and relation prediction on unseen text. Figure \ref{fig_leafai} shows a screenshot of the models correctly predicting entities and relations on an input sentence not present in the LCT corpus. As can be seen, the models are able to predict entities and relations with very high accuracy on new text, demonstrating the power of the corpus. 

\subsection*{Limitations}
\noindent The LCT corpus is designed as a granular and robust resource of annotated eligibility criteria to enable models for entity and relation prediction as means of query generation. The corpus does have a number of limitations however which should be recognized. First, the corpus is largely singly annotated, with 119 of 1,006 documents (11\%) double annotated and reconciled, while double annotation is generally considered to be the gold standard in the NLP research community. As discussed in the Technical Validation section, the reasonably high F\textsubscript{1} score from experiments to evaluate NER when training on the singly annotated portion of the corpus suggests relative consistency of annotation across both single and double annotated documents. Additionally, entities in roughly half of the LCT corpus (493 documents) were automatically predicted, then manually corrected. This can potentially lead to data bias if predicted entities are not thoroughly reviewed and corrected by human annotators. Similar results from our experiments to detect differences in performance by training on the manually annotated portion versus the semi-automatically annotated portion (F\textsubscript{1} scores of 78.6\% and 80.0\%) suggest this may not be not a significant issue. Last, though the ultimate goal of the LCT corpus is to facilitate more accurate query generation, the corpus itself is not composed of queries by which it can be compared to similar corpora and thus cannot necessarily be proven to be more effective. Similarly, as we do not formally define a quantifiable means for measuring semantic representation within annotations, it is difficult to demonstrate that the LCT corpus enables more accurate query generation. \\

\subsection*{Future Work}
\noindent As discussed, evaluation of generated query accuracy and semantic representation in annotations is difficult and can potentially be done by different methods, such as ROUGE scoring \cite{lin2004rouge} to compare generated query syntax to expected syntax, or by including UMLS Concept identifiers \cite{bodenreider2004unified} within the LCT annotation schema and comparing the number of UMLS concepts to those found in other corpora. \\

\noindent Taking a different approach, in future work, we intend to evaluate the LCT corpus and query generation methods by evaluating generated queries in the context of real clinical trials which have taken place at the University of Washington (UW). As the UW EHR system maintains clinical trial enrollments and patient identifiers alongside clinical data, it is possible to query our EHR databases to compare patients who actually enrolled in clinical trials versus those found by our queries had they been run at the time of a given trial. We believe this means of evaluation is uniquely valuable as it uses real world clinical trials and EHR data while scoring queries by the accuracy of their ultimate results rather than less consequential factors such as syntax.

\section*{Code availability}
\label{sec:code_availability}
\noindent All code used to generate, pre-annotate, and analyze the LCT corpus is freely available at \url{https://github.com/uw-bionlp/clinical-trials-gov-data}. \\

\noindent The LCT annotation guidelines can be found at \url{https://github.com/uw-bionlp/clinical-trials-gov-annotation/wiki}.

\section*{Acknowledgements} 

This study was supported in part by the National Library of Medicine under Award Number R15LM013209 and by the National Center for Advancing Translational Sciences of National Institutes of Health under Award Number UL1TR002319. Experiments were run on computational resources generously provided by the UW Department of Radiology.

\section*{Author contributions statement}

ND created the annotation guidelines, was primary annotator, and drafted the original manuscript. TM served as secondary annotator and reviewed and revised the manuscript. OU reviewed the task and goals as well as reviewed and revised the manuscript. MY conceived of the annotation task, supervised, and reviewed and revised the manuscript.

\section*{Competing interests}

The authors declare no competing interests.


\end{document}